\documentclass[journal]{IEEEtran}
\usepackage{amsmath}
\usepackage{amssymb}
\usepackage{graphicx}
\usepackage{booktabs}
 \usepackage{multirow}
 \usepackage{hyperref}
 \usepackage{xcolor}
 \usepackage{enumitem}
 \usepackage[english]{babel}
\usepackage[letterpaper,
            top=0.83in,
            bottom=0.6in,
            left=0.67in,
            right=0.67in,
            includeheadfoot]{geometry}
 
\ifCLASSINFOpdf
\else
\fi
\begin{document}
%
\title{DIFF-IPPO: Diffusion-Based Informative Path Planning with Open-Vocabulary Belief Maps}
%
%
%

\author{Sausar Karaf*, Oleg Sautenkov*, Mikhail Martynov, Dzmitry Tsetserukou
\thanks{*Equal contribution}
\thanks{Authors are with Intelligent Space Robotics Laboratory, CDE, Skoltech,
         Bolshoy Boulevard, 30, bld. 1, Moscow 121205, Russia
        {\tt\small {sausar.karaf, oleg.sautenkov, mikhail.martynov, d.tsetserukou} @skoltech.ru}}}

\maketitle

\begin{abstract}
Exploration and object search require robots to perceive their environment, identify regions of interest, and plan trajectories that improve target-detection likelihood or maximize information gain. Many IPP methods, especially in continuous environmental monitoring, rely on Gaussian-process belief models, while object-search settings often produce complex, multimodal belief maps from semantic or open-vocabulary perception. Global trajectory generation directly conditioned on such non-Gaussian belief maps remains comparatively underexplored. Although diffusion-based planners offer strong capabilities for modeling such distributions, their use in informative path planning remains limited. In this work, we propose \textit{DIFF-IPPO}, a pipeline that integrates an open-vocabulary belief map generator with a diffusion-based planner for global trajectory generation over belief maps. The method generates trajectories that concentrate sensor coverage over high-belief regions, achieving normalized detection scores between 81.49\% and 86.55\% across different dataset scenarios. We validate the system in a simulated search-and-rescue scenario where the planner searches candidate building regions to locate a burning building. In this setting, a team of five drones using batched belief-map-conditioned trajectory generation achieves first detections in 3.5 minutes.

\end{abstract}




%
\IEEEpeerreviewmaketitle

\section{Introduction}
%
%
%
%

\IEEEPARstart{A}{utonomous} exploration and object search are fundamental capabilities for robots operating in large, partially observable environments \cite{thrun2005probabilistic}. Informative path planning (IPP) addresses this problem by generating trajectories that maximize information gain, typically based on belief maps representing the spatial distribution of target likelihood \cite{bourgault2002information, marchant2014bayesian}.

Traditional IPP methods often rely on simplifying assumptions, such as Gaussian belief representations, and require computationally expensive optimization, limiting their applicability in real-time and multi-agent settings \cite{krause2008near, hollinger2014sampling}. Recent advances in generative models, particularly diffusion-based methods, enable efficient sampling from complex distributions and offer a promising alternative method for trajectory generation \cite{ho2020denoising, janner2022planning}. However, their application to informative path planning remains limited.

\begin{figure}[t!]
\centering
\includegraphics[width=0.85\columnwidth]{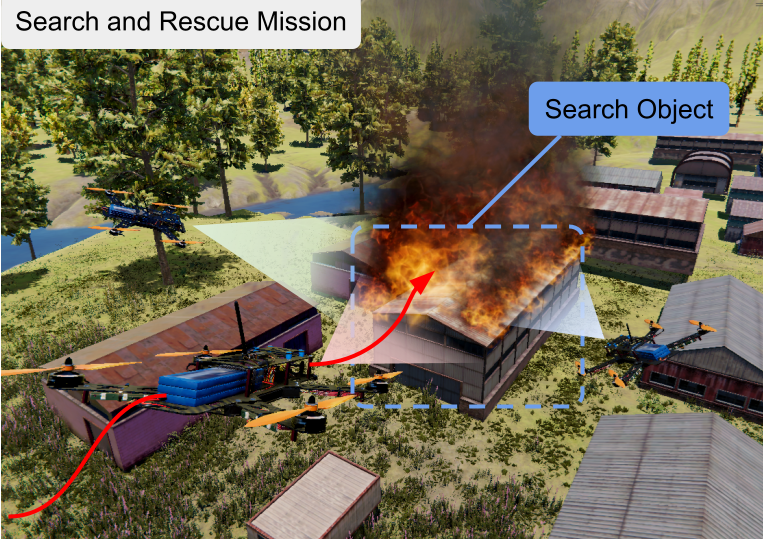}
\caption{Object search conducted by a team of drones in a scenario involving the detection of a burning building.}
\label{fig:intro_pic}
\end{figure}

\begin{figure*}[t!]
\centering
\includegraphics[width=0.85\textwidth]{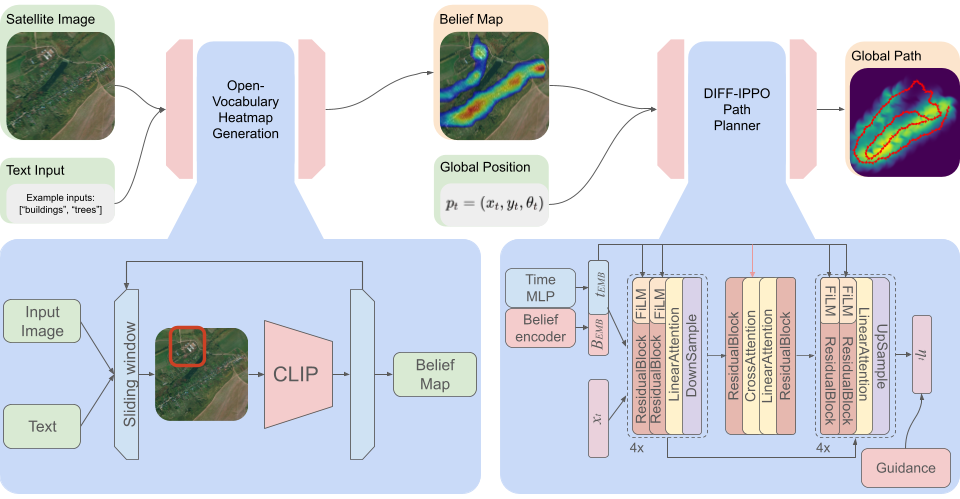}
\caption{The system architecture of the Diff-IPPO pipeline.}
\label{fig:system_architecture}
\end{figure*}

In this work, we propose \textit{DIFF-IPPO} (Fig.~\ref{fig:system_architecture}), a diffusion-based informative path planning framework that generates trajectories from belief maps. The pipeline consists of an open-vocabulary belief map generator that converts RGB satellite imagery and textual queries into belief maps using CLIP-based similarity~\cite{radford2021clip}, and a conditional diffusion model for trajectory generation. The framework is designed for trajectory generation over non-Gaussian belief maps and supports inference-time guidance through the coverage objective.

We evaluate \textit{DIFF-IPPO} on synthetic belief-map datasets (Fig.~\ref{fig:trajectory_grid}) and further evaluate the full pipeline in a simulated search-and-rescue environment. The synthetic experiments report normalized detection, path length, normalized coverage, and exploration efficiency across different belief-map distributions. We also evaluate batch trajectory generation in 3- and 5-trajectory settings, reporting total normalized detection, exploration efficiency, and redundancy.

The main \textbf{contributions} of this work are:
\begin{itemize}
    \item A diffusion-based informative path planning framework for trajectory generation over non-Gaussian belief maps.
    \item An integrated pipeline combining open-vocabulary belief map generation with diffusion-based trajectory generation.
    \item An inference-time guidance mechanism based on belief-map coverage.
    \item An evaluation on synthetic belief-map datasets and a simulated search-and-rescue environment.
\end{itemize}

\section{Related Work}

Informative path planning (IPP) has been widely studied for robotic exploration and object search. Classical approaches formulate IPP as an optimization problem that maximizes information gain over a belief map \cite{bourgault2002information, stachniss2005information}. Sampling-based methods, such as RRT-based planners, have been extended to incorporate information objectives, with algorithms such as TIGRIS \cite{moon2022tigris} and its incremental extension IA-TIGRIS \cite{moon2025iatigris}. 


Diffusion models have recently emerged as a powerful framework for trajectory generation due to their ability to model complex, multimodal distributions \cite{ho2020denoising}. Diffusion-based planning has been successfully applied in autonomous driving \cite{zheng2025diffusionbased}, quadruped locomotion \cite{Stamatopoulou2024DiPPeSTDP}, and multi-robot motion planning \cite{Shaoul2024MultiRobotMP}. These methods demonstrate strong capabilities in generating feasible and diverse trajectories while maintaining high performance in complex environments.

Several recent works have explored diffusion-based approaches specifically for multi-agent and swarm systems. SwarmDiffusion \cite{Zhura2025SwarmDiffusionET} and SwarmDiff \cite{Ding2025SwarmDiffSR} propose end-to-end diffusion frameworks for coordinating multiple agents in cluttered environments. Similarly, AID \cite{Lew2025AIDAI} introduces a diffusion-based approach for multi-agent informative path planning by modeling agent intent. While these methods show promising results, they either focus on general navigation or do not explicitly address the integration of perception-driven belief maps in the planning process.

In parallel, recent work has explored the integration of high-level semantic reasoning and language into robotic systems. Approaches such as FlockGPT~\cite{Lykov2024FlockGPTGU} and UAV-VLA~\cite{Sautenkov2025UAVVLAVS} leverage large language models and vision–language models, respectively, to guide multi-agent behavior. However, these methods primarily emphasize high-level coordination and do not explicitly address informative path planning or belief-driven exploration.

Our approach combines open-vocabulary perception with diffusion-based trajectory generation for informative path planning over non-Gaussian belief maps. The model learns a trajectory prior from expert demonstrations and supports the use of guidance during inference to influence trajectory generation. It also supports batched trajectory generation for producing multiple trajectories in parallel.

\section{Proposed Method}

\subsection{Problem Formulation}

We consider the problem of informative path planning for object search in a partially observable environment. Let $J(\boldsymbol{\tau})$ denote the expected information gain (or in our work expected detection reward) associated with a set of trajectories $\boldsymbol{\tau} = \{\tau^{(i)}\}_{i=1}^{M}$, where each trajectory $\tau^{(i)} = \{(x_t^{(i)}, y_t^{(i)}, \theta_t^{(i)})\}_{t=1}^{T}$ corresponds to the motion of agent $i$, with $(x_t^{(i)}, y_t^{(i)}, \theta_t^{(i)})$ representing its pose at time step $t$, and $T$ denoting the planning horizon. 

The environment is represented by a belief map $B \in \mathbb{R}^{H \times W}$, where $H$ and $W$ denote the height and width of the map, respectively. Each pixel in $B$ encodes the probability of the target object being present at the corresponding spatial location.

The objective is defined as maximizing the expected coverage of the belief map under trajectory-induced visibility:

\begin{equation}
\boldsymbol{\tau}^* = \arg\max_{\boldsymbol{\tau} \in \mathcal{T}^M} J(\boldsymbol{\tau}),
\end{equation}
where $\mathcal{T}^M$ denotes the space of feasible joint trajectories for all $M$ agents and $J(\boldsymbol{\tau})$ is defined in Section \ref{subsec:planner} as a function of belief map coverage under trajectory-induced visibility.

Collision avoidance is not explicitly modeled in this work, but can be incorporated during inference through additional guidance constraints or local planning.

\section{Open-Vocabulary Belief Map Generation}

The proposed method generates a belief map that localizes regions in an image corresponding to a given natural language query. This is achieved using a pretrained CLIP model, which jointly embeds images and text into a shared semantic space.

\text{1) Image Preprocessing:}  
The input image $I \in \mathbb{R}^{H \times W \times 3}$ is converted into a NumPy array and processed via a sliding-window method. A window of size $w \times w$ is moved across the image with stride $s$, producing overlapping crops.

\text{2) Text Encoding:}  
The input text query $q$ is encoded into a semantic feature vector using a pretrained CLIP text encoder:

\begin{equation}
\mathbf{f}_q =
\frac{\mathrm{Enc}_{\text{text}}(q)}
{\|\mathrm{Enc}_{\text{text}}(q)\|}
\end{equation}

Here, $\mathrm{Enc}_{\text{text}}(\cdot)$ denotes the CLIP text encoder, which maps a natural language query into a $d$-dimensional feature vector. The operator $\|\cdot\|$ represents the Euclidean ($L^2$) norm, used to normalize the feature vector to unit length. This normalization ensures that similarity computations depend only on the direction of the vectors (cosine similarity).

\text{3) Image Feature Extraction:}  
For each image crop centered at location $(x,y)$, a feature vector is extracted using the CLIP image encoder:

\begin{equation}
\mathbf{f}_{x,y} =
\frac{\mathrm{Enc}_{\text{img}}(I_{x,y})}
{\|\mathrm{Enc}_{\text{img}}(I_{x,y})\|}
\end{equation}

where $\mathrm{Enc}_{\text{img}}(\cdot)$ denotes the CLIP image encoder and $I_{x,y}$ is the cropped region.

\text{4) Similarity Computation:}  
The semantic similarity between the image crop and the text query is computed using the dot product between normalized feature vectors:

\begin{equation}
s_{x,y} = \mathbf{f}_{x,y}^\top \mathbf{f}_q
\end{equation}

Since both vectors are normalized, this corresponds to cosine similarity.

\text{5) Belief Map Aggregation:} Each similarity score $s_{x,y}$ is associated with a sliding window centered at $(x,y)$ and assigned to all pixels within that window. Because windows overlap, multiple scores contribute to each pixel location.

Let $\mathcal{W}_{i,j}$ denote the set of all window centers whose corresponding windows cover pixel $(i,j)$. The belief map is computed as:

\begin{equation}
\tilde{B}(i,j) = \sum_{(x,y)\in \mathcal{W}_{i,j}} s_{x,y}
\end{equation}

\begin{equation}
C(i,j) = |\mathcal{W}_{i,j}|
\end{equation}

\begin{equation}
B(i,j) = \frac{\tilde{B}(i,j)}{C(i,j)}
\end{equation}

This averaging ensures that each pixel reflects the mean similarity over all overlapping windows that include it.

\text{6) Normalization and Smoothing:}
The belief map is normalized to the range $[0,1]$ and smoothed using a Gaussian filter to reduce noise and improve spatial coherence.

\text{7) Thresholding:}  
To isolate regions with high semantic relevance, the belief map is filtered using a thresholding strategy (e.g., fixed threshold, percentile-based thresholding, or Otsu’s method), producing a sparse representation:

\begin{equation}
B_{\text{filtered}}(i,j) =
\begin{cases}
B(i,j), & \text{if } B(i,j) \geq \kappa \\
\text{NaN}, & \text{otherwise}
\end{cases}
\end{equation}
In the simulated environment experiments reported in this paper in Section \ref{subsec:sim_env}, we use percentile-based thresholding at the 80th percentile, retaining the top 20\% of belief-map responses.

\text{8) Visualization:} The final belief map is visualized by overlaying it on the original image using alpha blending.

Let $I_{\text{norm}} = I / 255$, and let $\phi(\cdot)$ denote a colormap mapping the belief map to an RGB representation.

\begin{equation}
I_{\text{overlay}} = (1 - \rho) I_{\text{norm}} + \rho \, \phi(B)
\end{equation}

where $I$ is the input image, $B \in [0,1]^{H \times W}$ is the normalized belief map, $\phi(B) \in [0,1]^{H \times W \times 3}$ is its colormap representation, and $\rho \in [0,1]$ controls blending strength.

\text{9) Grayscale and Binarized Representation:} For downstream tasks, a grayscale belief map is generated by resizing and scaling:

\begin{equation}
B_{\text{gray}} \in [0,255]^{n \times n}
\end{equation}

This grayscale map is further binarized to produce a binary mask:

\begin{equation}
B_{\text{bin}}(i,j) =
\begin{cases}
1, & \text{if } B_{\text{gray}}(i,j) \geq \tau \\
0, & \text{otherwise}
\end{cases}
\end{equation}

Additionally, a grayscale heatmap representation is generated for downstream tasks by resizing and scaling:
\begin{equation}
    H_{\text{bw}} \in [0,255]^{n \times n}
\end{equation}

The proposed open-vocabulary belief map generator uses a pretrained CLIP model to produce spatial belief maps from natural-language queries without task-specific training. In this work, these belief maps are used as inputs for object-search planning.

\subsection{Diff-IPPO Path Planner}
\label{subsec:planner}
Trajectory generation is performed using a conditional UNet-based diffusion model. At each diffusion timestep $k$, the network is conditioned on a spatial belief map, which is processed by a convolutional belief encoder to produce a set of belief tokens. These tokens are flattened, projected through a linear layer, and combined with a timestep embedding generated by a time MLP.
The resulting conditioning signals are injected into the UNet via residual temporal blocks with Feature-wise Linear Modulation (FiLM), enabling adaptive conditioning. The architecture follows an encoder--decoder structure with multi-scale processing: features are progressively downsampled using residual blocks with linear attention and subsequently upsampled through symmetric decoder blocks.
In addition, the belief tokens are incorporated at the bottleneck via cross-attention, allowing the model to integrate global contextual information from the belief map into the trajectory representation.

\subsubsection{Dataset}

To train the network, we constructed a synthetic dataset of belief maps paired with planner-generated trajectories. The belief maps consist of either irregularly shaped blobs or randomly scattered Gaussian distributions. For each belief map, we generated a trajectory using the simplified TIGRIS-inspired planner described in Section \ref{subsec:baselines}. The final dataset contains more than 42,000 samples, including 22,000 blob-based maps and 20,000 Gaussian-based maps.

\subsubsection{Network training}

The network is trained using a combination of diffusion, coverage, and smoothness objectives. The overall loss is defined as:
\begin{equation}
L = L_{\text{diff}} + L_{\text{cov}} + L_{\text{sm}}.
\end{equation}

The diffusion loss encourages accurate prediction of the noise added during the diffusion process and is given by:
\begin{equation}
L_{\text{diff}} =
\lambda_{\text{diff}}
\,
\mathcal{N}
\left(
\left\|
\hat{\epsilon} - \epsilon
\right\|_2^2
\right),
\end{equation}
where $\hat{\epsilon}$ and $\epsilon$ denote the predicted and ground-truth noise, respectively, $\lambda_{\text{diff}}$ is a scaling factor, and $\mathcal{N}(\cdot)$ denotes the frozen EMA normalized operator.

To evaluate coverage, we first model the visibility of the trajectory over the belief map. For each grid cell $(i,j)$ and time step $t$, we compute the distance from the agent to the cell center and the relative bearing angle. Let $(x_j, y_i)$ denote the spatial coordinates of the grid cell $(i,j)$. The distance $d_{ij}^{(t)}$ and the bearing angle $\phi_{ij}^{(t)}$ are defined as:
\begin{equation}
d_{ij}^{(t)} =
\sqrt{(x_j - x_t)^2 + (y_i - y_t)^2},
\end{equation}
\begin{equation}
\phi_{ij}^{(t)} =
\operatorname{atan2}(y_i - y_t,\; x_j - x_t).
\end{equation}

The relative viewing angle between the agent's heading $\theta_t$ and the direction to the grid cell is then given by:
\begin{equation}
\Delta\theta_{ij}^{(t)} =
\operatorname{atan2}
\left(
\sin(\phi_{ij}^{(t)} - \theta_t),
\cos(\phi_{ij}^{(t)} - \theta_t)
\right).
\end{equation}
where $\phi_{ij}^{(t)}$ denotes the absolute bearing from the agent's position $(x_t, y_t)$ to the grid cell $(x_j, y_i)$.

These quantities are used to define range and field-of-view attention terms:
\begin{equation}
R_{ij}^{(t)} =
\sigma
\left(
k_r (r_{\max} - d_{ij}^{(t)})
\right),
\end{equation}
\begin{equation}
A_{ij}^{(t)} =
\sigma
\left(
k_\theta(\theta_{\text{fov}} - |\Delta\theta_{ij}^{(t)}|)
\right),
\end{equation}
where $\sigma$ is the sigmoid function, $r_\text{max}$ is the maximum range, $\theta_{\text{fov}}$ is the field of view angle, and $k_r$ and $k_\theta$ are scaling coefficients.

These terms are combined to obtain the per-time-step visibility:
\begin{equation}
V_{ij}^{(t)} =
R_{ij}^{(t)} \cdot A_{ij}^{(t)}.
\end{equation}

The overall visibility across the trajectory is computed using a soft union:
\begin{equation}
V_{ij}
=
1 -
\prod_{t=1}^{T}
\left(
1 - V_{ij}^{(t)}
\right).
\label{eq:soft_union}
\end{equation}

This visibility is then used to estimate the portion of the belief map observed by the trajectory:
\begin{equation}
B_{ij}^{\text{vis}} =
B_{ij} \cdot V_{ij}.
\end{equation}

The total belief before and after observation is defined as:
\begin{equation}
T_{\text{before}} =
\sum_{i,j} B_{ij},
\end{equation}
\begin{equation}
T_{\text{after}} =
\sum_{i,j}
\left(
B_{ij} - B_{ij}^{\text{vis}}
\right).
\end{equation}

The coverage objective is defined as:
\begin{equation}
J(\boldsymbol{\tau}) =
1 -
\frac{
T_{\text{after}}
}{
T_{\text{before}} + \epsilon
}.
\label{eq:coverage_objective}
\end{equation}

The coverage loss penalizes the fraction of the belief map that remains unobserved:
\begin{equation}
L_{\text{cov}} =
\lambda_{\text{cov}}
\,\mathcal{N}
\left(
1 - J(\boldsymbol{\tau})
\right).
\label{eq:coverage_loss}
\end{equation}
where $\lambda_{\text{cov}}$ is a scaling factor for the coverage loss.

Finally, a smoothness loss is introduced to encourage dynamically feasible trajectories by penalizing large control inputs:
\begin{equation}
L_{\text{sm}} =
\lambda_{\text{sm}}
\mathcal{N}
\left(
\frac{1}{T}
\sum_{t=1}^{T}
\left(
w_j| j_t |^2 +
w_a| a_t |^2 +
w_v| v_t |^2
\right)
\right).
\end{equation}
where $v_t$, $a_t$, and $j_t$ denote the velocity, acceleration, and jerk at time step $t$ along the trajectory, respectively, $w_j$, $w_a$, and $w_v$ are scaling coefficients, and $\lambda_{\text{sm}}$ is a scaling factor for the smoothness loss.
The coverage term is also used as a guidance signal during inference to steer trajectory generation toward higher coverage regions.

\subsubsection{Batch Trajectory Generation}

The network generates multiple trajectories in parallel as a batch conditioned on the same belief map. The coverage objective is evaluated over the entire set of trajectories by aggregating their induced visibility using the visibility model defined above.

During inference, this objective is used as a guidance signal to steer the generation process toward trajectories that improve overall coverage under the shared belief map.

\subsection{Simulated Environment}

\begin{figure}[h]
\centering
\includegraphics[width=0.85\columnwidth]{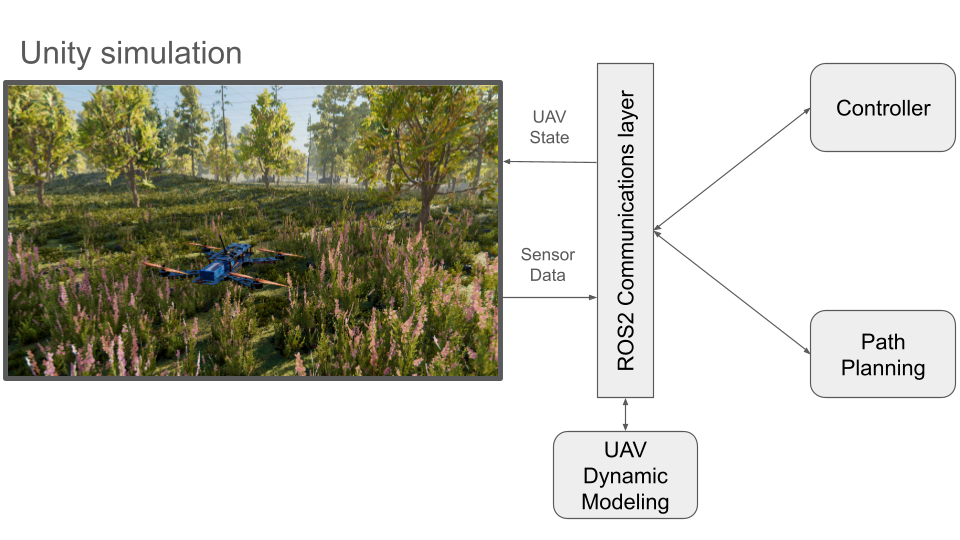}
\caption{Schematic representation of the simulated environment used in this study.}
\label{fig:simulation_scheme}
\end{figure}

To evaluate the developed algorithms, we constructed a simulated environment inspired by the Flightmare simulator \cite{song2020flightmare}. The simulator was implemented using the Unity 3D game engine \cite{unity}. A depth camera is simulated within the environment using UnitySensors \cite{unitysensors}. The drone dynamics are computed by a separate ROS2 node. Control of the drone is also handled via dedicated ROS2 nodes for the controller and the path planner. A schematic representation of the simulator architecture is shown in Fig. \ref{fig:simulation_scheme}.

The drone dynamics used in this work are defined as follows:
\begin{equation}
\dot{x} =
\begin{bmatrix}
v \\
g + \frac{1}{m} R^{T} e_3 \, T \\
W(\phi, \theta)\,\omega \\
J^{-1} \left( \tau - \omega \times (J \omega) \right)
\end{bmatrix}
\end{equation}
\noindent
where $x = [p^T, v^T, \eta^T, \omega^T]^T \in \mathbb{R}^{12}$ is the state, with position $p \in \mathbb{R}^3$, velocity $v \in \mathbb{R}^3$, Euler angles $\eta = [\phi,\theta,\psi]^T$, and body angular velocity $\omega \in \mathbb{R}^3$. The constant $m$ denotes the mass, $g \in \mathbb{R}^3$ is gravity, $R \in SO(3)$ is the rotation matrix, and $e_3 = [0,0,1]^T$. The inertia matrix is $J \in \mathbb{R}^{3 \times 3}$, $T \in \mathbb{R}$ is the total thrust, and $\tau \in \mathbb{R}^3$ is the control torque. The matrix $W(\phi,\theta)$ maps body angular velocity to Euler angle rates, and $\omega \times (J\omega)$ denotes the gyroscopic term.

The controller is based on a formulation similar to that presented in \cite{Mellinger}.
\begin{equation}
f_{\mathrm{des}} = -K_p e_p - K_v e_v + m (g + a_d),
\end{equation}
\begin{equation}
u_1 = f_{\mathrm{des}}^\top R e_3,
\end{equation}
\begin{equation}
\tau = -K_R e_R - K_\omega e_\omega,
\end{equation}
\begin{equation}
u = K^{-1} \begin{bmatrix} u_1 \\ \tau \end{bmatrix}
\end{equation}
\noindent
where $K_p, K_v, K_R, K_\omega \in \mathbb{R}^{3 \times 3}$ are gain matrices, $e_R \in \mathbb{R}^3$ and $e_\omega \in \mathbb{R}^3$ denote attitude and angular velocity errors, respectively, $u_1 \in \mathbb{R}$ is the collective thrust, $\tau \in \mathbb{R}^3$ is the control torque, $u \in \mathbb{R}^4$ is the motor command vector, and $K \in \mathbb{R}^{4 \times 4}$ is the control allocation matrix.

\section{Experiments and Results}
\subsection{Baseline Comparison}
\label{subsec:baselines}
\begin{table*}[t!]
\caption{Performance Comparison Across Datasets}
\label{tab:dataset_comparison}
\centering
\begin{tabular*}{\textwidth}{@{\extracolsep{\fill}}llccccc@{}}
\toprule
\textbf{Dataset} & \textbf{Method}  & \textbf{Norm. Det. ↑} & \textbf{Path Len. ↓} & \textbf{Norm. Cov. ↓} & \textbf{Expl. Eff.  $\cdot 10^4$↑} \\
\midrule

\multirow{4}{*}{All} 
& Diff IPPO   & \textbf{0.8149} & 842.72  & 0.3917 & \textbf{9.87} \\
& AID (Belief map input)   & 0.6667 & 746.49  & 0.3790 & 8.92 \\
& TIGRIS-inspired planner    & \textbf{0.8189} & 841.57  & 0.3090 & \textbf{9.95} \\

\midrule

\multirow{4}{*}{Irregular Blobs} 
& Diff IPPO    & \textbf{0.8595} & 910.75  & 0.4106 & 9.52 \\
& AID (Belief map input)  & 0.7669 & 750.61  & 0.3693 & \textbf{10.22} \\
& TIGRIS-inspired planner   & 0.7412 & 868.81  & 0.3339 & 8.65 \\

\midrule

\multirow{4}{*}{Scattered Gaussians} 
& Diff IPPO    & \textbf{0.8655} & 745.02  & 0.3425 & \textbf{12.24} \\
& AID (Belief map input)     & 0.6701 & 742.63  & 0.3437 & 9.02 \\
& TIGRIS-inspired planner    & \textbf{0.8881} & 617.98  & 0.2364 & \textbf{15.19} \\

\bottomrule

\end{tabular*}
\end{table*}

\begin{figure}[t]
    \centering
    \includegraphics[width=0.85\columnwidth]{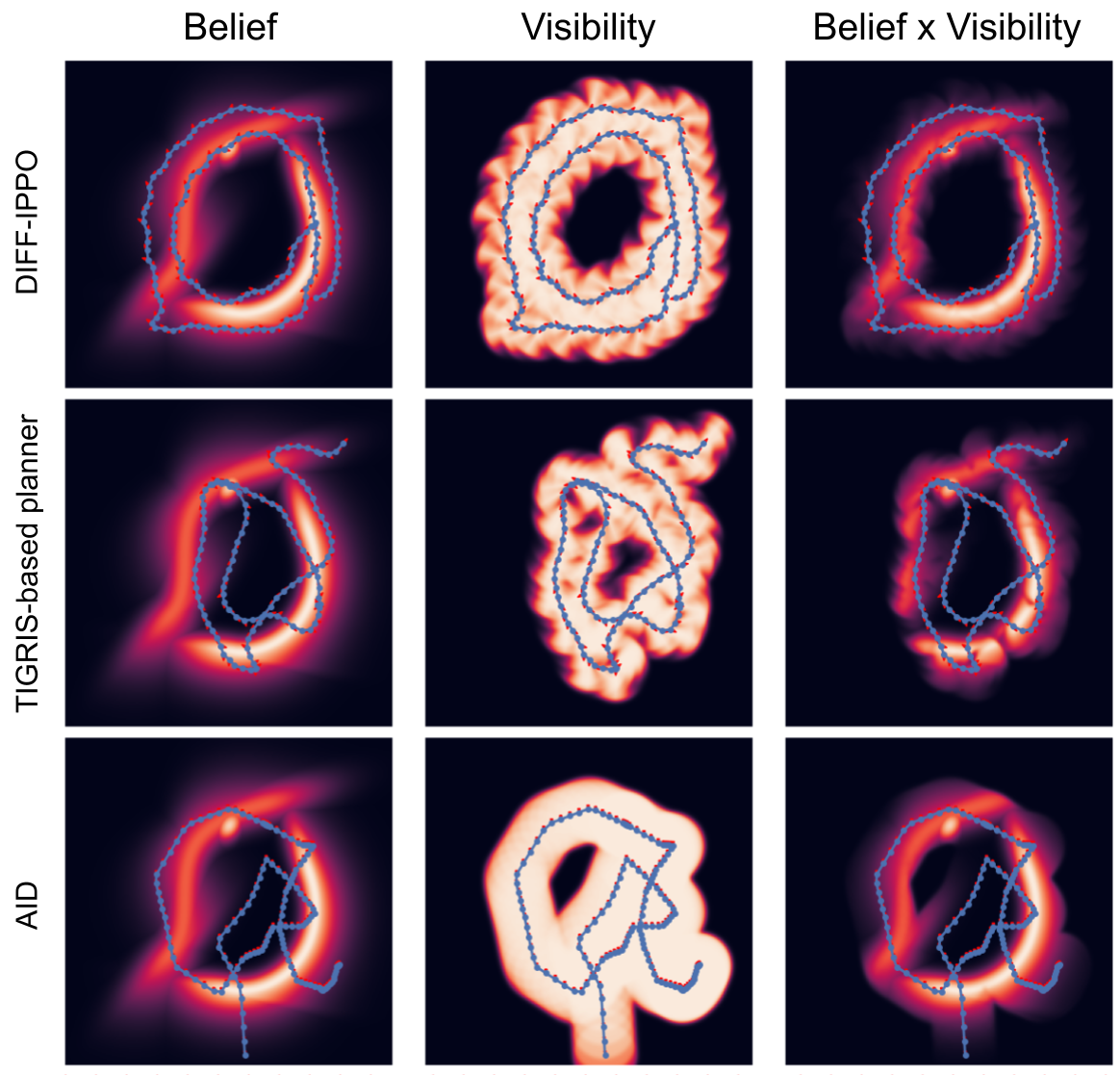}
    \caption{Example trajectories generated by the planners considered in this study. The first column shows the belief map. The second column shows the visibility map, illustrating what the agent observes when following the trajectory. The third column depicts the observed belief map after incorporating observations.}
    \label{fig:trajectory_grid}
\end{figure}

We evaluate the performance of Diff-IPPO using synthetically generated belief maps. To assess its effectiveness, we measure path length and compute the expected detection probability (Eq. \ref{eq:expected_detection}), total coverage (Eq. \ref{eq:coverage}), and exploration efficiency (Eq. \ref{eq:efficiency}) achieved by the generated trajectories. These metrics are then compared against those of several representative informative path planning methods. Fig. \ref{fig:trajectory_grid} illustrates example trajectories generated by the planners considered in this study.
\begin{equation}
P_{det} = \frac{\sum_{i,j} B(i,j)\, V(i,j)}{\sum_{i,j} B(i,j)};
\label{eq:expected_detection}
\end{equation}
\begin{equation}
C = \frac{1}{HW} \sum_{i,j} V(i,j);
\label{eq:coverage}
\end{equation}
\begin{equation}
\eta = \frac{P_{det}}{L},
\label{eq:efficiency}
\end{equation}
where $H$ and $W$ are the height and width of the belief map respectively, $L$ is the length of the trajectory.

In this work, we compare our method with two informative path planning baselines: AID and a simplified TIGRIS-inspired planner.

AID \cite{Lew2025AIDAI} is used as a diffusion-based baseline. Since our task uses a belief map rather than a Gaussian-process uncertainty map, we provide the belief map through AID's uncertainty input. This biases AID toward regions with larger belief values instead of uniform map coverage. Because AID does not explicitly model agent heading, visibility is evaluated using an isotropic $360^\circ$ field of view, which is an optimistic assumption compared with heading-aware directional sensing.

We also use a simplified Python implementation of a TIGRIS-inspired planner based on Moon et al. \cite{moon2022tigris}. The planner retains the high-level ideas of belief-informed sampling, budgeted tree construction, optimistic Bayesian entropy-reduction rewards, and path-integrated sensing. However, it uses straight-line planar motion, a two-dimensional sector sensor with constant detection rates, discretized edge observations, fixed-iteration planning, and batch replanning from scratch. The original Near, pruning, and closed-node operations are not implemented.

The experiments were conducted on three groups of datasets. The first group consists of randomly generated irregularly shaped blobs. The second group contains sparse Gaussian distributions scattered across the map, and the third group combines characteristics from both. Table~\ref{tab:dataset_comparison} summarizes the results of these experiments.

All planners were implemented within a unified Python framework. Diff-IPPO achieves normalized detection scores of 81.49\%, 85.95\%, and 86.55\% on the All, Irregular Blobs, and Scattered Gaussians datasets, respectively. The corresponding exploration efficiency values are 9.87, 9.52, and 12.24 ($\cdot 10^4$). Path lengths for Diff-IPPO are 842.72, 910.75, and 745.02 across the three datasets, while normalized coverage values are 0.3917, 0.4106, and 0.3425.

\subsection{Multi-Trajectory Evaluation}

For these experiments, the planner generates multiple trajectories in parallel as a batch conditioned on the same belief map. The induced visibility for each trajectory $b$ is denoted by $V_b(i,j)$.

To quantify overlapping coverage, we define the following aggregation terms. The union visibility is given by:
\begin{equation}
V_{\mathrm{union}}(i,j) =
1 - \prod_{b=1}^{K} \big(1 - V_b(i,j)\big)
\end{equation}
where $K$ denotes the number of trajectories in the batch.

The total accumulated visibility is defined as:
\begin{equation}
V_{\mathrm{total}} = \sum_{b=1}^{K} \sum_{i,j} V_b(i,j).
\end{equation}

Redundant visibility is computed as:
\begin{equation}
V_{\mathrm{redundant}} = \sum_{i,j} \left( \sum_{b=1}^{K} V_b(i,j) - V_{\mathrm{union}}(i,j) \right).
\label{eq:redundant_vis}
\end{equation}

The redundancy ratio is defined as:
\begin{equation}
R = \frac{V_{\mathrm{redundant}}}{V_{\mathrm{total}}}.
\label{eq:redundancy}
\end{equation}

\begin{figure}[t]
    \centering
    \includegraphics[width=1\columnwidth]{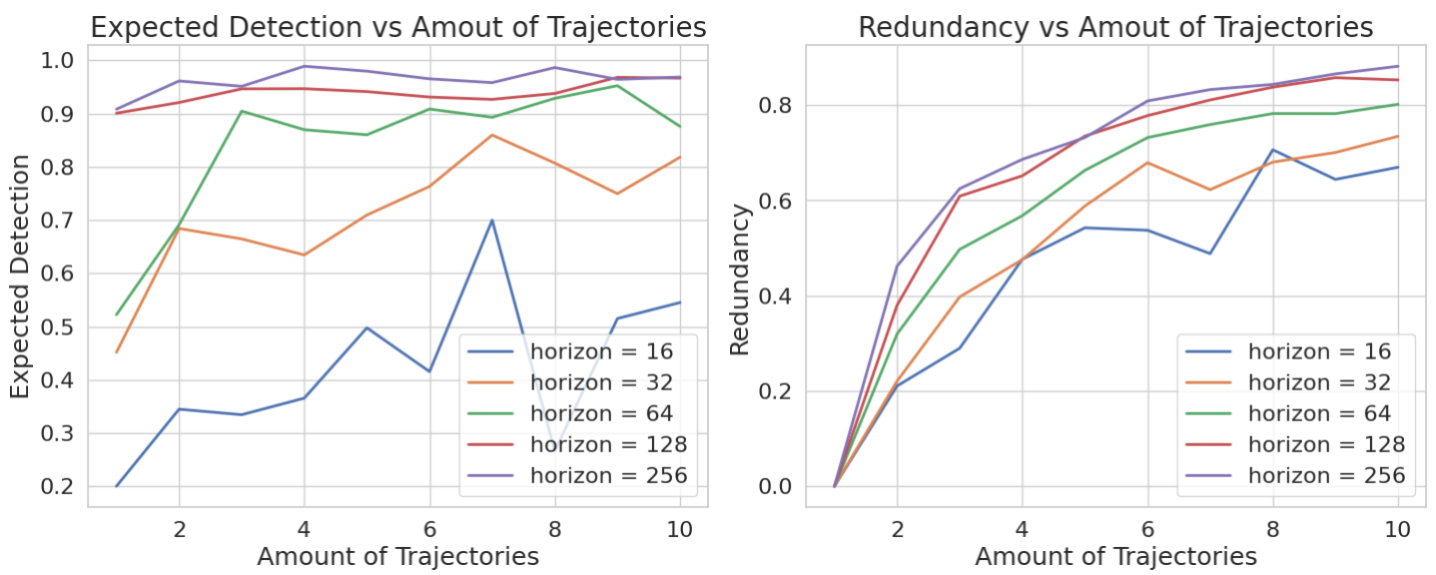}
    \caption{Graph depicting the expected detection (\textit{left}) and redundancy (\textit{right}) as a function of the number of trajectories for different planning horizon values.}
    \label{fig:horizon_vs_amount}
\end{figure}

\begin{figure}[t]
    \centering
    \includegraphics[width=0.5\columnwidth]{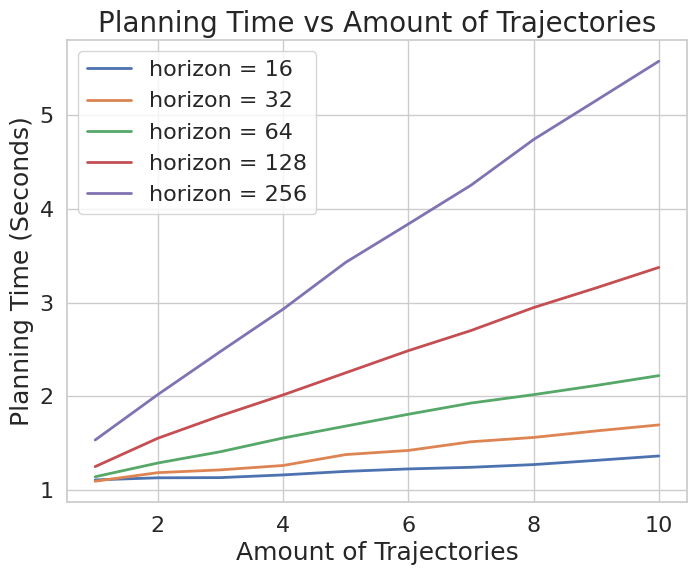}
    \caption{Graph depicting planning time as a function of the number of trajectories for different planning horizon values.}
    \label{fig:multi_agent_time}
\end{figure}

Figure~\ref{fig:horizon_vs_amount} reports expected detection and redundancy as functions of the number of generated trajectories for different planning horizon values. In the reported experiments, expected detection generally increases with both trajectory horizon and the number of trajectories, with smaller changes observed beyond a horizon of 64 and more than four trajectories. Redundancy also increases with the number of trajectories across the evaluated horizon values. Figure~\ref{fig:multi_agent_time} reports planning time as a function of the number of trajectories and planning horizon.

\begin{figure}[t]
    \centering
    \includegraphics[width=0.85\columnwidth]{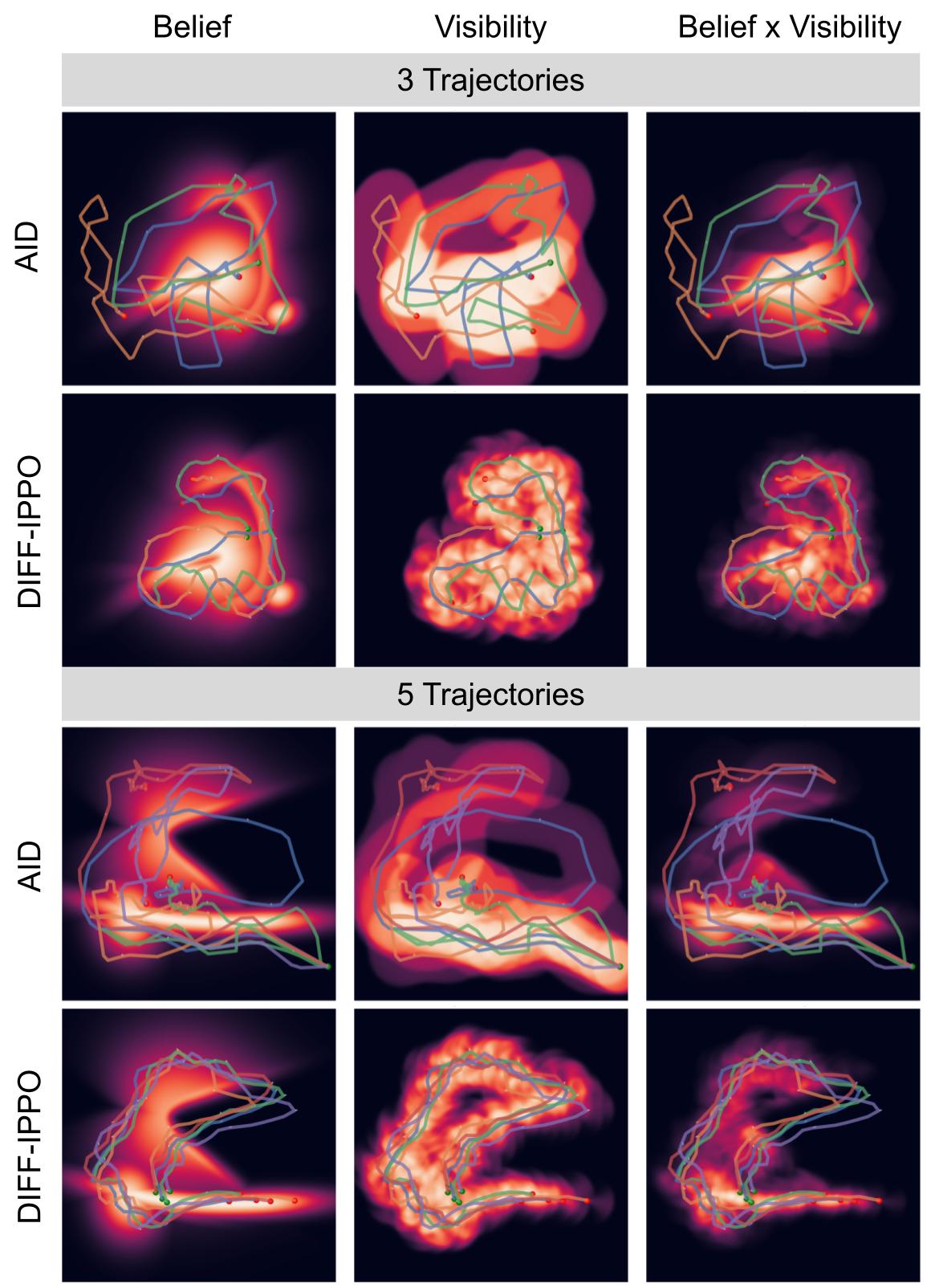}
    \caption{Example trajectories generated by the Diff-IPPO and AID planners for 3- and 5-trajectory settings on the irregular blob dataset.}
    \label{fig:ma_traj_grid}
\end{figure}

Diff-IPPO was compared to the AID planner. Example trajectories are shown in Figure~\ref{fig:ma_traj_grid}, and the results are summarized in Table~\ref{tab:drone_redundancy}. Redundancy values differ across settings: Diff-IPPO reports lower redundancy in the 3-trajectory case, while AID reports lower redundancy in the 5-trajectory case.

\begin{table*}[t]
\caption{Batch Trajectory Evaluation: 3 vs. 5 Trajectories}
\label{tab:drone_redundancy}
\centering
\begin{tabular*}{\textwidth}{@{\extracolsep{\fill}}llcccccc@{}}
\toprule
\textbf{Config} & \textbf{Method}  & \textbf{Avg. Norm. Det. ↑} & \textbf{Avg. Path Len. ↓} & \textbf{Norm. Det. ↑} & \textbf{Redundancy ↓} & \textbf{Expl. Eff. $\cdot 10^4$↑} \\
\midrule

\multirow{2}{*}{3 Trajectories} 
& AID         & \textbf{0.6301} & 742.0931 & 0.8402 & 0.4139 & 8.49 \\
& Diff IPPO    & 0.5750 & \textbf{471.6409} & \textbf{0.9852} & \textbf{0.4006} & \textbf{12.56} \\
\midrule

\multirow{2}{*}{5 Trajectories} 
& AID        & 0.5510 & 738.5147 & 0.9078 & \textbf{0.5622} & 7.46 \\
& Diff IPPO   & \textbf{0.5692} & \textbf{500.7105} & \textbf{0.9896} & 0.6229 & \textbf{11.63} \\
\bottomrule

\end{tabular*}
\end{table*}

\subsection{Simulated Environment Experiment}
\label{subsec:sim_env}
\begin{figure}[t!]
    \centering
    \includegraphics[width=0.85\columnwidth]{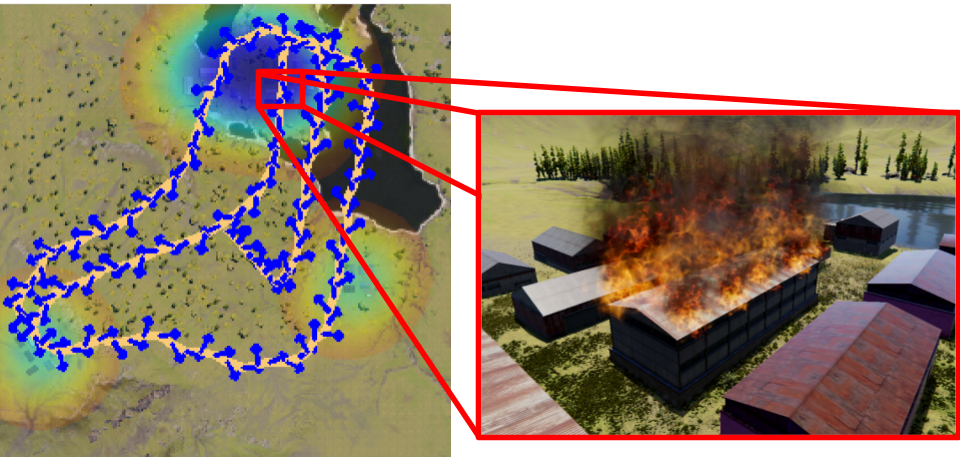}
    \caption{Example of a trajectory generated for a single drone performing area search for a burning building in the simulated environment. The image on the left shows the simulated satellite view of the environment, along with the generated belief map produced using the query “building” and the corresponding planned trajectory. The red box indicates the actual location of the burning building.}
    \label{fig:sim_example}
\end{figure}

The Diff-IPPO pipeline was also evaluated in the simulated environment (Fig. \ref{fig:sim_example}). An open-vocabulary belief map generator, described previously, was used to convert simulated RGB satellite images into belief maps. The objective was to locate a burning building within a $1\,\mathrm{km} \times 1\,\mathrm{km}$ environment. The global planner outputs planar waypoints $(x,y,\theta)$. For execution in the 3D simulator, these waypoints are lifted to fixed-altitude setpoints $(x,y,z_0,\theta)$, where $z_0$ is kept constant. The low-level controller tracks these setpoints using the UAV dynamics model. An operator monitored the camera feeds from the drones, and a run was considered successful once the burning building was clearly visible.

This scenario was evaluated using one, three, and five drones, with five runs conducted for each configuration. The building was randomly selected from one of three towns. Diff-IPPO successfully found the target in all cases, with average first detection times of 7.9 minutes for a single drone, 4.25 minutes for three drones, and 3.5 minutes for five drones.

\section{Conclusion and Future Work}

In this work, we presented \textit{DIFF-IPPO}, a diffusion-based informative path planning pipeline that integrates an open-vocabulary belief map generator with a diffusion model for trajectory generation over belief maps. The proposed framework is designed for planning over non-Gaussian belief representations derived from natural language queries and visual inputs.

We evaluated the method across multiple belief map types and report normalized detection scores ranging from 81.49\% to 86.55\% across different dataset scenarios. The complete pipeline was also validated in a simulated search-and-rescue scenario for fire detection in a $1\,\mathrm{km} \times 1\,\mathrm{km}$ environment. In this setting, the system was used to guide multiple drone trajectories for target search, resulting in successful detections within the simulated runs.

The multi-trajectory experiments in this work should be interpreted as batched trajectory generation conditioned on a shared belief map. The model is not trained on joint multi-agent demonstrations and does not include explicit task allocation, communication, inter-agent collision avoidance, or decentralized coordination. Therefore, the present work does not constitute a full multi-agent informative path planning framework, but rather a step toward belief-map-conditioned batch trajectory generation.

Future work includes improving performance on sparse and highly irregular belief maps, further investigating trajectory diversity in batch generation settings, and extending \textit{DIFF-IPPO} toward full multi-agent informative path planning. Additional extensions may explore full 3D trajectory generation, explicit safety-aware planning, inter-agent coordination, and reasoning modules for belief map refinement.


%





\ifCLASSOPTIONcaptionsoff
  \newpage
\fi



%



\bibliographystyle{ieeetr}
\bibliography{bibtex/bib/references}

\vfill

\end{document}